%% file: sig_main.tex
\documentclass[sigconf]{acmart}

\AtBeginDocument{%
  \providecommand\BibTeX{{%
    \normalfont B\kern-0.5em{\scshape i\kern-0.25em b}\kern-0.8em\TeX}}}

\setcopyright{acmcopyright}
\copyrightyear{2018}
\acmYear{2018}
\acmDOI{10.1145/1122445.1122456}

\acmConference[SIGIR '21]{SIGIR '21:ACM SIGIR Conference on Research and Development in Information Retrieval}{}{}
\acmBooktitle{SIGIR '21:ACM SIGIR Conference on Research and Development in Information Retrieval}
\acmPrice{15.00}
\acmISBN{978-1-4503-XXXX-X/18/06}

\usepackage[nodayofweek,level]{datetime}
\RequirePackage{adjustbox}
\RequirePackage{soul}
\RequirePackage[normalem]{ulem}
\RequirePackage{url}
\usepackage{multirow}
\usepackage{subcaption}

\input{math_commands}

\usepackage{hyperref} 
\usepackage{paralist}
\RequirePackage{adjustbox}

\RequirePackage[inline]{enumitem} \setlist{nosep}
\setlist[description]{leftmargin=.5em}
\usepackage[textsize=scriptsize]{todonotes}
\dbltextfloatsep \textfloatsep
\intextsep \textfloatsep
\makeatletter
\g@addto@macro{\normalsize}{%
\setlength{\abovedisplayskip}{3pt plus1pt}%
\setlength{\abovedisplayshortskip}{3pt plus1pt}%
\setlength{\belowdisplayskip}{3pt plus1pt}%
\setlength{\belowdisplayshortskip}{3pt plus1pt}}

\makeatother

\def\shortname{S3\xspace}
\def\longname{Select, Substitute, Search\xspace}
\def\model{\shortname{}\xspace}
\def\data{{\mbox{\textsc{OKVQA}}\textsubscript{\shortname}}\xspace}
\def\newdata{\shortname{}{VQA}\xspace}
\def\dnametype1{\data\xspace}
\def\dnameother{OKVQA$\setminus$\shortname\xspace}
\def\mrcsingle{{\mbox{\textsc{MRC}}\textsubscript{sing}}\xspace}
\def\mrcmultiple{{\mbox{\textsc{MRC}}\textsubscript{mult}}\xspace}
\def\topten{{\mbox{T10}\textsubscript{text}}\xspace}

\begin{document}

\title{\longname: A New Benchmark for\\
Knowledge-Augmented Visual Question Answering}

\author{Aman Jain}
\authornote{Both authors contributed equally to this research.}
\author{Mayank Kothyari}
\authornotemark[1]
\affiliation{%
    \institution{IIT~Bombay}
    \country{India}
    }
\author{Vishwajeet Kumar}
\affiliation{%
    \institution{IBM Research, India}
    \country{}
    }
\author{Preethi Jyothi}
\affiliation{%
    \institution{IIT~Bombay}
    \country{India}
    }
\author{Ganesh Ramakrishnan}
\affiliation{%
    \institution{IIT~Bombay}
    \country{India}
    }
\author{Soumen Chakrabarti}
\affiliation{%
    \institution{IIT~Bombay}
    \country{India}
    }

\renewcommand{\shortauthors}{Aman and Mayank, et al.}
\begin{abstract}
   Multimodal IR, spanning text corpus, knowledge graph and images, called outside knowledge visual question answering (OKVQA), is of much recent interest.  However, the popular data set has serious limitations. A surprisingly large fraction of queries do not assess the ability to integrate cross-modal information. Instead, some are independent of the image, some depend on speculation, some require OCR or are otherwise answerable from the image alone. To add to the above limitations, frequency-based guessing is very effective because of (unintended) widespread answer overlaps between train and test folds.  Overall, it is hard to determine when state-of-the-art systems exploit these weaknesses rather than really infer the answers, because they are opaque and their `reasoning' process is uninterpretable. An equally important limitation is that the dataset is designed for the quantitative assessment only of the end-to-end answer retrieval task, with no provision for assessing the correct (semantic) interpretation of the input query. In response, we identify a key structural idiom in OKVQA, {\em viz.}, \shortname{} ({\em select}, {\em substitute} and {\em search}), and build a new data set and challenge around it.  Specifically, the questioner identifies an entity in the image and asks a question involving that entity which can be answered only by consulting a knowledge graph or corpus passage mentioning the entity.  Our challenge consists of (i)\data, a subset of OKVQA annotated based on the structural idiom and (ii)\newdata, a new dataset built from scratch.   We also present a neural but structurally transparent OKVQA system, \model, that explicitly addresses our challenge data set, and outperforms recent competitive baselines. We make our code and data available at \url{https://s3vqa.github.io/}
\end{abstract}

\maketitle


\input{main/sec1-introduction}
\input{main/sec2-limitations}

\input{main/sec3-new-dataset}
\input{main/sec4-prior}
\input{main/sec5-proposed}
\input{main/sec6-exp-results}
\input{main/sec7-conclusion}

\bibliographystyle{ACM-Reference-Format}

\bibliography{references, voila, anthology}  
\end{document}

%% file: math_commands.tex
\usepackage{amsthm,amsmath,amsfonts,bm,xspace}
\usepackage{bbm}
\usepackage{color}

\def\eqref#1{(\ref{#1})}

\def\1{\bm{1}}

\DeclareMathAlphabet{\mathsfit}{\encodingdefault}{\sfdefault}{m}{sl}
\SetMathAlphabet{\mathsfit}{bold}{\encodingdefault}{\sfdefault}{bx}{n}



%% file: main/sec1-introduction.tex
\section{Introduction} \label{sec:intro} 

\begin{figure}[th]
    \centering
    \includegraphics[width=.5\hsize]{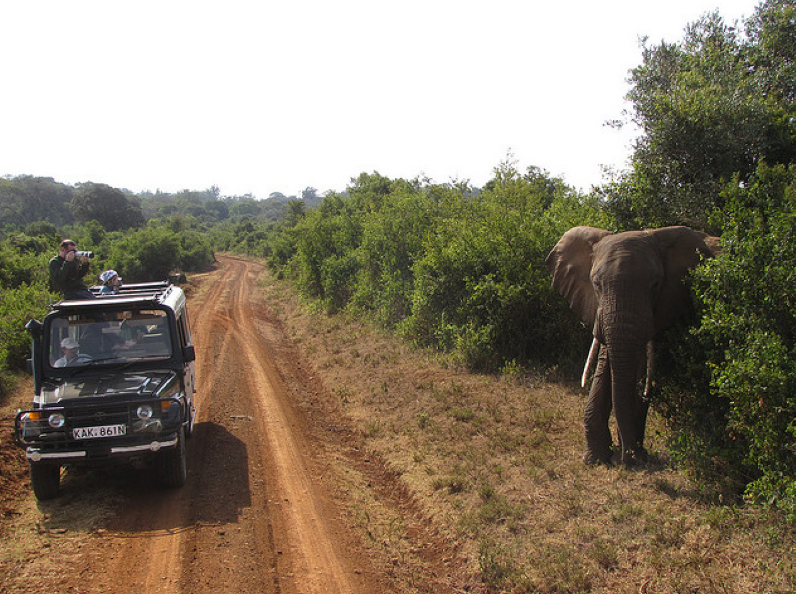}
    \caption{Sample OKVQA image and query: `What is {\itshape this animal} poached for?'. As discussed in Section~\ref{sec:newDataChallenge}, we associate the following additional ground truth with each such question: `{\em What is this <select, substitute=``elephant''>animal</select> poached for?}'}
    \label{fig:okvqa_example}
\end{figure}

Multimodal question answering (QA), specifically, combining visual and textual information, is of much recent interest \citep{antol2015vqa, krishna2017visualgenome, hudson2019gqa, malinowski2014multiworld}.
In the simplest visual QA (VQA) variant, the query is textual and a given image suffices, by itself, to answer it.
A more challenging variant is where outside knowledge (OK) from a corpus or knowledge graph needs to be combined with image information to answer the query~\citep{marino2019okvqa}.
This task is referred to as OKVQA.

An example OKVQA instance is shown in Figure~\ref{fig:okvqa_example}.
The system cannot answer the question by detecting objects from the image, or by recognizing colour, count, shape as in traditional VQA~\citep{antol2015vqa, krishna2017visualgenome, hudson2019gqa, malinowski2014multiworld}). The system first needs to understand the relationship between the image content and the question and then needs to query the outside world for retrieving relevant facts (in this particular case, about the poaching of elephants).

For our purposes, a host of modern image processing techniques~\citep{rcnn, fasterRcnn} (which we consider as black boxes) can identify image patches representing salient entities (elephant, man, jeep, tree, sky, camera) and some spatial relations between them ({\em e.g.}, man holding camera, man sitting in a jeep, {\em etc.}).  This is called a \emph{scene graph}.  Knowledge that elephant is-a animal may come from a knowledge graph~\cite{BollackerEPST2008Freebase} or open information extraction~\citep{Etzioni2011FCSMOpenIe2}.  Treated thus, OKVQA is less of a visual recognition challenge and more of a challenge to fuse unstructured text and (graph) structured data extracted from the image, which is relevant to the Information Retrieval (IR) community.

\subsection{Discontent with current state of OKVQA}
\label{sec:okvqa}

\paragraph{Small step vs.\ giant leap}
    QA in the TREC community \citep{Voorhees1999trecqa, Voorhees2001trecqa} began as ``corpus QA'' long before neural NLP became widespread.  Naturally, any structural query interpretation or scoring of responses was completely transparent and deliberate.  QA entered the modern NLP community as ``KGQA'' \cite{BerantCFL2013SEMPRE, YaoVD2014Jacana}, which was focused on precise semantic interpretation of queries.
    Initial efforts at corpus QA in the NLP community also had modest goals: given a query and passage, identify a span from the passage that best answers the query.  Such ``reading comprehension'' (RC) generalized proximity-sensitive scoring~\citep{LvZ2009positional, ButtcherCL2006ProximityBm25, PetkovaBC2007NamedEntityProximity}.
    Compared to these ``small steps'', OKVQA was a ``giant leap'', placing enormous faith in opaque neural machinery to answer visual queries that have no controlled space of reasoning structure. Some questions require optical character recognition to `read' street signs in the image, others are speculative (``can you guess the celebration where people are enjoying?''), questions that are independent of the image (``Pigeons are also known as what?''), and questions that require only the image and no outside information to answer (``Which breed of dog is it?'').
    
\paragraph{The unwelcome power of guesswork}
    Rather than score candidate answers from a large set, such as entities in a KG, or spans from passages in a large open-domain corpus, current systems treat OKVQA as a classification problem over the $k$ most frequent answers in the training set.  We found a  huge train-test overlap, which condones and even rewards this restrictive strategy - 48.9\% answers in the test set of OKVQA~\cite{marino2019okvqa} were found to be present in the train set.

\paragraph{Retrospection}
    Such limitations are not specific to OKVQA, but have been witnessed in other communities as well.  Guessing has been found unduly successful for the widely used SQuAD QA dataset~\citep{rajpurkar-etal-2016-squad, clark-etal-2019-dont}.  The complex QA benchmark HotPotQA \citep{Xiao+2019DFGNhotpotqa} was supposed to test multi-step reasoning.  However, \citet{min-etal-2019-compositional} found that top-scoring QA systems as well as humans could guess the answer without such reasoning for a large fraction of queries.  Their conclusion is particularly instructive: ``there should be an increasing focus on the role of evidence in multi-hop reasoning and possibly even a shift towards information retrieval style evaluations with large and diverse evidence collections''.  Even more damning is the work of \citet{tang2020multihop}, who report that ``state-of-the-art multi-hop QA models fail to answer 50--60\% of sub-questions, although the corresponding multi-hop questions are correctly answered''.  Other tasks such as entailment \citep{mccoy-etal-2019-right} have faced similar fate.
    
\paragraph{Remedies}
    Awareness of the above-mentioned pitfalls have led to algorithmic improvements, principally by reverting toward explicit query structure decomposition.  Training data can be augmented with decomposed clauses \citep{tang2020multihop}.  The system can learn to decompose queries by looking up a single-clause query collection \citep{zhang-etal-2019-complex, perez-etal-2020-unsupervised}.  Explicit and partly interpretable control can be retained on the assembly of responses for subqueries~\citep{sun-etal-2018-open, sun-etal-2019-pullnet}.
    Earlier datasets and evaluations are being fixed and augmented as well~\citep{clark-etal-2019-dont, clark2018think, ribeiro-etal-2019-red}.
    New datasets are being designed to be compositional from the ground up~\citep{talmor-berant-2019-multiqa}, so that system query decomposition quality can be directly assessed.
    Here we pursue both the data and algorithm angles:
    we present a new data set where the structural intent of every query is clear by design, and we also propose an interpretable neural architecture to handle this form of reasoning. 
    Our new dataset and our supporting algorithm are designed keeping in mind the interpretability of the entire system and {\bf can thus potentially contribute to advancement of research in other applications that include question answering over video, audio as well as over knowledge graphs}. As mentioned earlier, most state-of-the-art VQA systems are trained to fit an answer distribution; this  brings into question their ability to generalize well to test queries. To address this issue and improve the explanatory capabilities of VQA systems, prior work has investigated the use of visual explanations~\cite{das2017human} and textual explanations~\cite{park2018multimodal, wu:aaai21} to guide the VQA process. Our new challenge dataset and our proposed OKVQA architecture has the advantage of being inherently explanatory, without having to rely on any external explanations. {\bf Further, we contribute a new benchmark \newdata\ that is explanatory in its very genesis, as will be described in Section~\ref{sec:newdata}.}

\subsection{A new OKVQA Challenge Dataset}
\label{sec:newChallenge}
    
    In response to the serious shortcomings we have described above, we present a new OKVQA challenge. Our dataset consists of two parts, each of which is provided with rich, compositional annotations that provision for assessment of a system's query reformulation
    {\em viz.}, (i)  \data, a subset of the OKVQA~\cite{marino2019okvqa} dataset with such compositional annotations (Section~\ref{sec:okvqasubset}) and (ii) \newdata, a completely new dataset ({\em c.f.}, Section~\ref{sec:newdata}) that has been sanitized from the ground up against information leakage or guesswork-friendly answering strategies.  
    Every question in our new benchmark is guaranteed to require integration of KG, image-extracted information, and search in an open-domain corpus (such as the Web).
    Specifically, the new data set focuses on a specific but extremely common reasoning idiom, shown in Figure~\ref{fig:okvqa_example}.  The query pertains to an entity $e$ in the image (such as elephant), but refers to it in more general terms, such as type $t$ (animal) of which $e$ is an instance.  There is no control over the vocabulary for mentioning $e$ or~$t$. In our running example, an OKVQA system may implement a \emph{substitution} to generate the query ``what is this animal poached for?'', drawing `elephant' out of a visual object recognizer's output. While many reasoning paradigms may be important for OKVQA, we argue that this is an important one.  Our data set exercises OKVQA systems in fairly transparent ways and reveals how (well) they perform this basic form of reasoning.  Addressing other structures of reasoning is left for future work. 
    Despite this restricted reasoning paradigm, which ought to be well within the capabilities of general-purpose state-of-the-art OKVQA systems, we are surprised to find existing OKVQA models yield close to $0$ evaluation score on \newdata.

\subsection{An interpretable OKVQA system}
\label{sec:interpretableOKVQA}
    
    Continuing in the spirit of ``small steps before giant leap'', we present \model{} ({\em c.f.}, Section~\ref{sec:Proposal}), a neural OKVQA system that targets this class of queries and reasoning structure.  Our system thus has a known interpretation target for each query, and is therefore interpretable and affords systematic debugging of its modules.  \model{} has access to the query and the scene graph of the image.  It first {\em selects} a query span to {\em substitute} with (the string description of) an object from the scene graph; this object, too, has to be selected from many objects in the image.  Sometimes, the question span and the object description would be related through an instance-of or subtype-of relation in a typical KG or linguistic database; ideally, \model{} should take advantage of this signal.  There are now two ways to use the reformulated question.  We can regard answer selection as a classification problem like much of prior work, or we can prepare a Web search query, get responses, and learn to extract and report answer span/s in the RC style (referred to as the open-domain setting).  \model{} is wired to do both ({\em c.f.}, Section~\ref{sec:ExptResults}).

%% file: main/sec2-limitations.tex
\section{Limitations in existing OKVQA data}
\label{sec:OldData}

\begin{figure}
    \centering
    \begin{subfigure}[b]{0.2\textwidth}
        \centering
        \includegraphics[width=\textwidth]{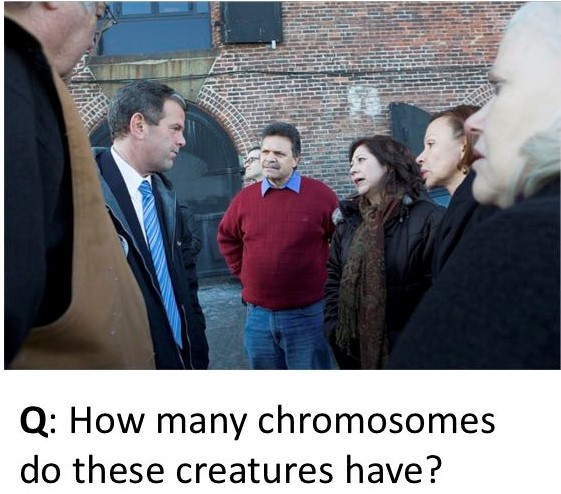}
        \caption[Type 1]%
        {{\small Type 1}}    
        \label{fig:type1}
    \end{subfigure}
    \hfill
    \begin{subfigure}[b]{0.2\textwidth}  
        \centering 
        \includegraphics[width=\textwidth]{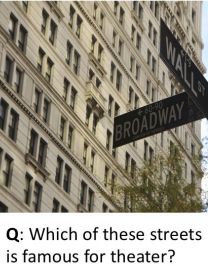}
        \caption[]%
        {{\small Type 2}}    
        \label{fig:type2}
    \end{subfigure}
    \vskip\baselineskip
    \begin{subfigure}[b]{0.2\textwidth}   
        \centering
        \includegraphics[width=\textwidth]{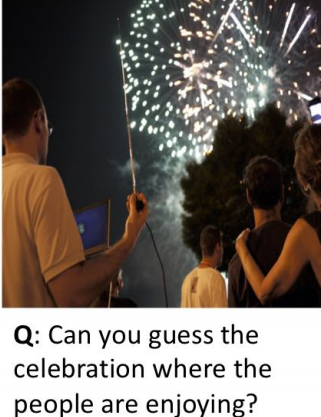}
        \caption[]%
        {{\small Type 3}}    
        \label{fig:type3}
    \end{subfigure}
    \hfill
    \begin{subfigure}[b]{0.2\textwidth}   
        \centering 
        \includegraphics[width=\textwidth]{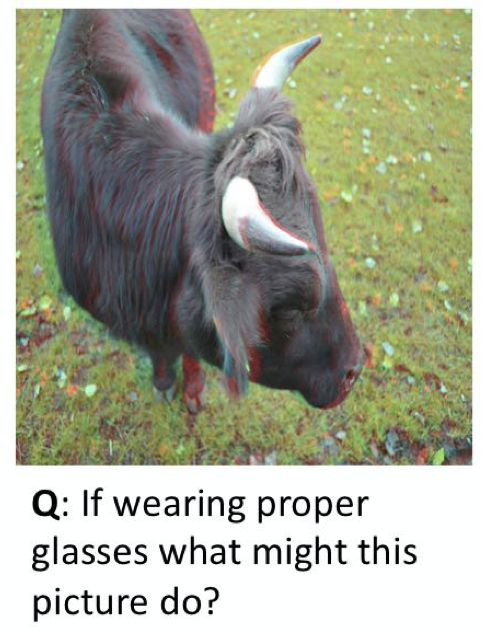}
        \caption[]%
        {{\small rest}}    
        \label{fig:rest}
    \end{subfigure}
    \caption[Example question for each category]
    {{\small Example question for each category}}
    \label{fig:cat_of_quest}
\end{figure}

The widely used benchmark OKVQA dataset \citep{marino2019okvqa} consists of over 14000 question-image pairs, with 9000 training examples and 5000 examples in the test set.  We identify two broad issues with it.

First, significant overlap exists between answers in the train and test folds. Recall from Section~\ref{sec:okvqa} that 48.9\% of answers in the test set are present in the training set. Existing systems leverage this limitation to boost their accuracy by limiting their test answers to the most frequent answers in the training set.

Second, unlike WebQuestions~\citep{berant-etal-2013-semantic} or ComplexWebQuestions~\citep{talmor-berant-2018-web}, OKVQA questions, even when grouped into some categories (see below) have no clear pattern of reasoning.  18\% (\textbf{type-1}) of the questions require detecting objects and subsequent reasoning over an external knowledge source to arrive at the answer.  7\% of the questions (\textbf{type-2}) require reading text from the image (OCR) (and no  other information) 
to answer.  12\% of the questions (\textbf{type-3}) are based on personal opinion or speculation.
The remaining questions (\textbf{rest}) can perhaps be described best through Figure~\ref{fig:rest}.  We provide an example of each type in Figure~\ref{fig:cat_of_quest}.

We found that several queries of \textbf{type-1} have a structural similarity to the bridging queries in Complex\-Web\-Questions \citep{talmor-berant-2018-web}. There, each query has exactly two clauses.  The first clause, when issued to a Web search engine, returns (via RC) an entity which plugs into the second clause, which is again sent to the Web search engine, fetching the overall answer.  We found that \textbf{type-1} questions can be reformulated, with the help of the scene graph, to a query that can be answered directly using Web search.  Inspired by Complex\-Web\-Questions, we next develop our challenge data set.

%% file: main/sec3-new-dataset.tex
\section{Design of a new OKVQA Challenge data set}
\label{sec:newDataChallenge}

Consider the \textbf{type-1} query associated with Figure~\ref{fig:okvqa_example}: ``what is \uwave{this animal} poached for?''  We discovered that such questions can be answered through a sequence of three well-defined steps:
\begin{description}[leftmargin=.5em]
    \item[Select:] This step identifies a span in the question (``this animal'') that needs to be replaced with some information from the image, as part of the query reformulation process.  As part of our dataset(s), we make available the ground truth span for each question, to facilitate the development and evaluation of the selection model,  independent from other components of a OKVQA system.  We hypothesize that any OKVQA system that fetches the correct answer must solve this subproblem correctly in the first place.
    
    \item[Substitute:]
    Having identified the query span to replace, the {\em substitute} operation determines a key-phrase associated with the image that should replace the selected span. Such a key-phrase could be the name of an object in the image ({\em e.g.}, `elephant') or some attributes within the image (such as color), or a relationship between objects/people ({\em e.g.}, `carrying in hand'). As part of our dataset(s), we release the keyphrase substituting each selected span in the question.  This data can be used for independent training and evaluation of implementations of the substitution operation, assuming oracle assistance from the other operations.

    \item[Search:] After the query has been reformulated as described above, it can be used to harvest candidate answers from a corpus or a KG. The reformulated query in Figure~\ref{fig:okvqa_example} will be ``what is \uwave{elephant} poached for?'' with associated ground truth answer `tusk'. By providing gold query reformulations and gold answers, we also facilitate the evaluation of the {\em search} operation, independent of the selection and substitution operations.
\end{description}

As another example, the question in Figure~\ref{fig:type1} can be reformulated from ``how many chromosomes do \uwave{these creatures} have'' to ``how many chromosomes do \uwave{humans} have?'' by replacing the hypernym or super-type ``these creatures'' with the hyponym or sub-type `humans'.  Again, the reformulated query can be answered using an open domain QA system.  

The following two subsections, we describe the two parts of our challenge data set.  First, in Section~\ref{sec:okvqasubset}, we describe \textbf{\data}, produced by subsetting and annotating OKVQA to fit the specifications justified above.  Next, in Section~\ref{sec:newdata}, we describe \textbf{\newdata}, created from the ground up to our specifications.

\subsection{\data: Annotated subset of OKVQA}
\label{sec:okvqasubset}

    This is a subset of the OKVQA dataset in which we have annotated every question with spans, substitutions and gold answers. 
    More specifically, we divide the OKVQA dataset into two parts based on the category of the question. As discussed in Section~\ref{sec:OldData}, the first part consists of all the questions of category \textbf{Type~1} and the second part consists of rest of the questions. We refer to the first part as \dnametype1 and the second part as \dnameother. The name \dnametype1 represents three operations (Select, Substitute and Search) needed to answer all the questions of category \textbf{Type1}. \dnameother is the remaining OKVQA dataset after removing \dnametype1.

    2640 of the 14000 question-image pairs in OKVQA are of {\bf type-1}, amounting to 18.8\% of the total. For these questions, our ground truth annotations also include the object and span to be used for select and substitution for every image question pair. The average length of the span selected was 2.44 words (13.14 characters), with a standard deviation of 1.36 (7.24). When COCO~\cite{coco}, ImageNet~\cite{imagenet} and OpenImages \citep{openimages} object detection models were run on this set, 24.7\% of the question-image pairs did not have the ground truth object in the detections owing to detection errors and vocabulary limitations.

\subsection{\newdata: New dataset built from scratch} 
\label{sec:newdata}

    Gathering experience from the process of annotating the subset \data\ of OKVQA, we build a new benchmark dataset \newdata\ in a bottom-up manner, pivoting on the {\em select}, {\em substitute} and {\em search} operations, applied in that order, to each query. Our \newdata{} data is built on the Open Images collection~\cite{openimages}. For each entity/object/class name $e$ ({\em e.g.}, peacock) in the OpenImages dataset, we identified its corresponding `parent' label $t$ ({\em e.g.}, bird) that generalizes~$e$. We employed the hierarchical structure specified within the Open Images dataset itself to ensure that $e\to t$ enjoy a hyponym-hypernym or instance-category relation.
    
    Next, we (semi-automatically) identified an appropriate Wikipedia page for $e$ and, using the Wikimedia parser~\cite{wikimedia} on that page, extracted text snippets. A question generation (QG) model based on T5 \citep{T5} was used to generate $<$ {\em question}, {\em answer}$>$ pairs from these snippets. We retained only those pairs in which the {\em question} had an explicit mention of~$e$. Note that unlike OKVQA, by design, the \newdata\ dataset has exactly one correct ground truth {\em answer} per {\em question}. In our experimental results (in Section~\ref{sec:ExptResults}, specifically Table~\ref{table:opendomain}) we observe how this leads to a much stricter evaluation of search systems and therefore much lower numbers for \newdata\ in comparison to OKVQA, which offers 10 answers per question~\citep{marino2019okvqa}.
    
    Subsequently, each mention of $e$ in the filtered question set was replaced with the `parent' label $t$ using manually defined templates. One such template is to replace $e$ and the determiner preceding it ({\em e.g.} a, an, the) with the string - `this $t$'. Finally, this question set was filtered and cleaned up  manually  using the following guidelines: (i)~any question which could be answered just by using the image or was not fact-based was eliminated; (ii)~any question that needed grammatical improvements was corrected, without changing its underlying meaning.
    
    We also provided paraphrased questions (generated using T5, and Pegasus\cite{zhang2020pegasus}) as suggestions along with each templated question for the annotators to pick instead of the original templated question in order to bring in variety among the questions. Finally, for each question we pick an image from OpenImages corresponding to the object $e$ referred in it. We ensure that we pick a distinct image each time in case the object $e$ is referred in multiple questions. 

%% file: main/sec4-prior.tex
\section{Prior architectures}
\label{sec:PriorSystems}

VQA that requires reasoning over information from external knowledge sources has recently gained a lot of research interest. VQA systems~\citep{marino2019okvqa} have started incorporating external knowledge for question answering. Existing methods use one of the following approaches to integrate external knowledge:

\begin{enumerate}[leftmargin=*]
    \item Retrieve relevant facts about objects in the image and entities in the question and reason over extracted facts to arrive at the final answer for the question. 
    \item Collect evidence from text snippets returned from a search engine and extract answer from the text snippet. 
\end{enumerate}

The baseline architecture proposed by~\citet{marino2019okvqa} introduced ArticleNet to retrieve Wikipedia articles relevant to the question and image entities. ArticleNet encodes a Wikipedia article using a GRU and it is trained to predict whether the ground truth answer is present in the article. The hidden states of sentences in the article are used as the encoded representation. This encoded representation is given as input to a multimodal fusion model along with the question and image features. The answer is predicted by formulating the problem as a classification on the most frequent answers in the training set, leveraging the huge overlap in answers in training and test sets. \citet{garderes-etal-2020-conceptbert}~jointly learn knowledge, visual and language embeddings. They use ConceptNet \citep{Speer_Chin_Havasi_2017} as the knowledge source, and graph convolution networks~\citep{kipf2016semi} to integrate the information.  Similar to the OKVQA baseline system, they also formulate the problem as classification on the most frequent answers in the training set~\citep{yu2020crossmodal}.

%% file: main/sec5-proposed.tex
\section{Proposed architecture}\label{sec:poloss} %
    \label{sec:Proposal}
    
    \begin{figure*}[t!]
    \includegraphics[width=.95\linewidth]{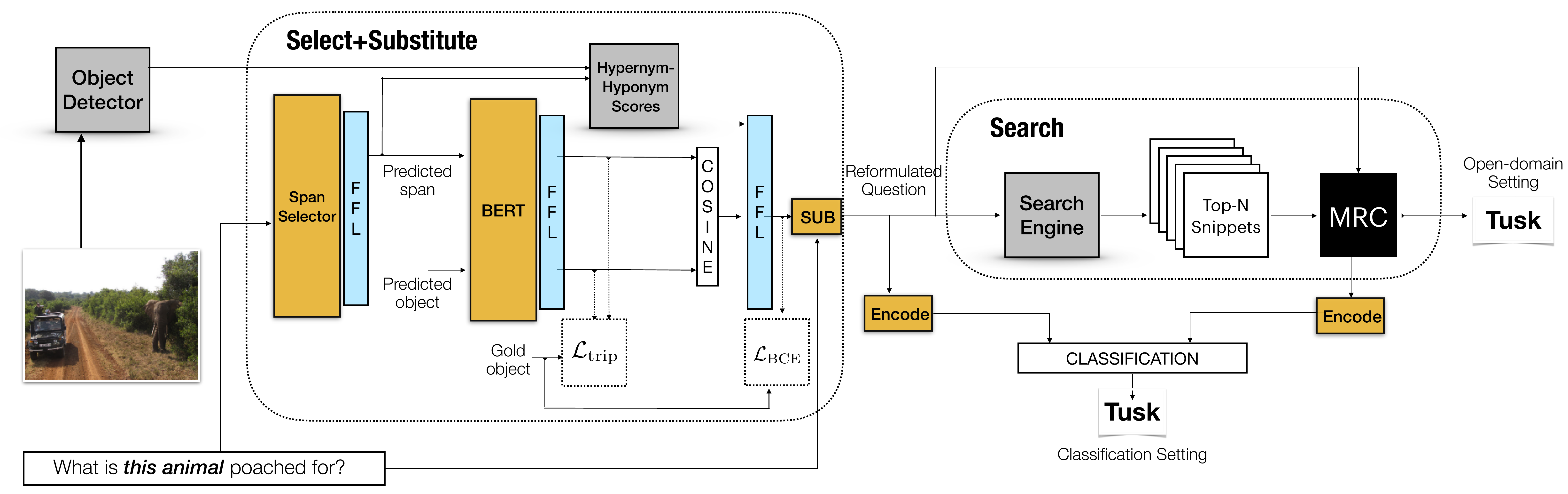}
    \centering
    \caption{Schematic diagram of \model. FFL refers to feedforward layers that are trained using our datasets.}
    \label{fig:arch}
    \end{figure*}

    Figure~\ref{fig:arch} shows an overall schematic diagram of our proposed \shortname{} (\longname) architecture. As described in Section~\ref{sec:newDataChallenge}, \textit{Select} and \textit{Substitute} are two operations that, when sequentially applied to a question, result in a reformulated query that could be answered without further reference to the image. Question reformulation is followed by a \textit{Search} operation that involves a Machine Reading Comprehension (MRC) module to extract answers from the top snippets retrieved via a search engine for each of the reformulated questions.
    
    \subsection{Select and Substitute} 
    \label{select&substitute}
    The \textit{Select} operation entails determining a span from the question that serves as a good placeholder for an object in the image. We refer to this module as the \textbf{SpanSelector} in Figure~\ref{fig:arch}. To implement this module, we use a large pre-trained language model (such as BERT~\cite{devlin2018bert}) with additional feed-forward layers (denoted by FFL1) that are trained to predict the start and end of the span. These additional layers are trained to minimize the cross-entropy loss between the predicted spans and the ground truth spans that accompany all the questions in our dataset.
    
    Once we have a predicted span from the SpanSelector, we want to determine which object from the image is best described by the question span. First, we rely on a state-of-the-art object detection system to provide a list of most likely objects present in the image. Next, we aim to identify one object from among this list that would be most appropriate as a substitution for the span that the question is centred around. This is achieved by minimizing the following combined loss function that predicts an object as the best substitute for a question span:
\begin{align}
\mathcal{L} = \alpha \cdot \mathcal{L}_{\text{trip}}(s,p,n) + (1 - \alpha) \cdot \mathcal{L}_{\text{BCE}}(s,p,n)
\end{align}
where $\mathcal{L}_{\text{trip}}$ is a triplet loss defined over triplets of embeddings corresponding to the question span ($s$), the ground-truth (gold) object ($p$) and a detected object different from the ground-truth object ($n$), $\mathcal{L}_{\text{BCE}}$ is a binary cross-entropy loss over each detected object and the span with the gold object acting as the reference, and $\alpha$ is a mixing coefficient that is tuned as a hyperparameter. Here, $\mathcal{L}_{\text{trip}}$ is defined as:
\begin{align}
\mathcal{L}_{\text{trip}}(s,d,p) = \max(d(s,p) - d(s,n) + m, 0)
\end{align}
where $d(\cdot,\cdot)$ is a distance function on the embedding space (such as cosine distance) and $m$ is a predefined margin hyperparameter that forces examples from the same class ($s$ and $p$) to be closer than examples from different classes ($s$ and $n$). (Both $\mathcal{L}_{\text{trip}}$ and $\mathcal{L}_{\text{BCE}}$ have been explicitly listed in Figure~\ref{fig:arch}.)

Computing $\mathcal{L}_{\text{BCE}}(s,p,n)$ also involves the use of a hypernym-hyponym scorer that examines whether or not each (question span, detected object) pair exhibits a hypernym-hyponym relationship. The scorer makes use of a lexical database such as WordNet~\cite{wordnet} that explicitly contains hypernym-hyponym relationships between words.
We map the span and detected object to the corresponding synsets in WordNet, and use the pre-trained Poincar\'e embeddings by~\citet{poincare} to embed them. These pre-trained embeddings were trained on the (hypernymy based) hierarchical representations  of the word sequences in WordNet synsets by embedding them into an $n$-dimensional Poincar\'e ball by leveraging the distance property of hyperbolic spaces. 

Finally, the reformulated question is obtained by substituting the predicted span in the question with the object that is predicted with highest probability as being the best substitute for the span.

\subsection{Search} 

After the question has been reformulated, we pass it through Google's search engine to retrieve the top 10 most relevant snippets.%
\footnote{To ensure reproducibility, we release these snippets as part of our dataset.}
These snippets are further passed as input to a Machine Reading Comprehension (MRC) module. An MRC system takes a question and a passage as its input and predicts a span from the passage that is most likely to be the answer. Using the MRC module, we aim at meaningfully pruning the relatively large amounts of text in Google's top 10 snippets and distilling it down to what is most relevant to the reformulated question.

The MRC module also allows us to move beyond the classification setting (as mentioned in Section~\ref{sec:okvqa}) and directly predict an answer given a question and its corresponding context. We refer to this as the open-domain task, shown in Figure~\ref{fig:arch}. For fair comparisons to prior work that adopt the classification setting, we also support classification as shown in Figure~\ref{fig:arch}. Here, embeddings for both the reformulated question and the output from the MRC are concatenated and fed as input to a softmax output layer over the top $k$ answers in the training set.

%% file: main/sec6-exp-results.tex
\section{Experiments and Results}
\label{sec:ExptSettings}

We report  experiments with \data and \newdata, using our \model\ model in comparison with a strong baseline (the BLOCK~\cite{ben2019block} model). In the following subsections, we explain the approaches being compared (Section~\ref{sec:methods}), evaluation settings employed in the experiments (Section~\ref{sec:eval}) and the results (Section~\ref{sec:ExptResults}).

\subsection{Methods}
\label{sec:methods}

\textbf{BLOCK~\cite{ben2019block}: } BLOCK is a state-of-the-art multimodal fusion technique for VQA, where the image and question pairs are embedded in a bimodal space. This bimodal representation is then passed to a softmax output layer, to predict the final answer. Note that, as per the currently prevalent practice in OKVQA (referred to in Sections~\ref{sec:okvqa} and~\ref{sec:PriorSystems}) the softmax is performed over the $k$ most frequent 
answers in the training set. We chose BLOCK as our baseline, since it is an improved version of a VQA system called MUTAN~\cite{MUTAN} that was originally reported as a baseline for the OKVQA~\cite{marino2019okvqa} dataset. By design, BLOCK is a vanilla VQA model that takes an image and question as its inputs and predicts an answer. Modifying it to additionally accept context as an input did not help improve its accuracies. 

\noindent \textbf{\model: } As described in Section~\ref{sec:Proposal}, our model \model{} processes each question using the {\em select}, {\em substitute} and {\em search} operations. 

We use SpanBERT~\cite{joshi-etal-2020-spanbert} as our MRC system to extract candidate answers from the relevant text snippets. We use the pretrained BERT-base model%
\footnote{\url{https://github.com/google-research/bert}}
to implement the select and substitute operations. 

\subsection{Evaluation} 
\label{sec:eval}

\begin{description}
    \item[Closed-domain (classification) setting:]  In this setting, the answer is predicted from a predefined vocabulary, that is constructed using the top-$k$  answers in the training set. Since \data{} is a relatively small subset of the OKVQA dataset and since all systems were observed to improve in their evaluation scores with increasing values of $k$, we report numbers by setting $k=1084$ in all experiments, which corresponds to setting the predefined answer vocabulary to the entire train set. The evaluation criteria for \data is the standard VQA metric~\cite{VQA} which is defined as:
    \begin{align}
            \mathrm{Accuracy}(\text{ans}) = min \Big\{\frac{\# humans\ that\ said\ \text{ans}}{3}, 1\Big\}
    \end{align}
    For \newdata, we note that accuracy is calculated by doing an exact string match with answer since we have exactly one correct ground truth answer per question.
    In retrospect, 42.1\% of answers in the test set of \data are found to be present in the train set. Thus 42.1\% serves as a generous skyline in the classification setting for \data\ (for \newdata, the overlap is 11.6\%). 
    \item[Open-domain setting:]  The vocabulary of answers is unconstrained while predicting the answer. The predicted answer should exactly match one of the annotated answers to be counted as correct. The evaluation metric is the same as in the classification setting.
\end{description}

\subsection{Implementation details} 

\begin{description}
    \item \textbf{Select module} To select the span, the input question is first encoded using a pretrained BERT model. BERT outputs a 768 dimensional representation for each of the tokens in the input. Each token's output representation is then passed to a linear layer (size:$768 \times 2$) with two values in the output that correspond to the probability of the token being the start (and end) of the desired span. This linear layer is trained with cross entropy loss using the gold span labels. During prediction, the tokens with the highest values for start and end are used to mark the start and end of the spans. In case the start token comes after the end token during prediction, an empty span is returned.
    
    \item \textbf{Substitute module} It comprises two parts: a) We apply a pretrained BERT model to encode the input question and all the other reformulated questions using each detection as the replacement for the span. BERT outputs a 768 dimensional representation for each of the tokens in the input. We compute the representation for a span using the averaged representation of all its tokens. The span representation and all other BERT representations of the detections within the reformulated questions are passed through a linear layer (size: $768 \times 2048$) followed by $\mathrm{tanh}$ activation and another linear layer (size: $2048 \times 1024$). We use the span as the anchor, the gold object as the positive instance and all other detections as negative instances for the triplet loss. 
    b) We take the representation of the span and the detection from the network above and compute the cosine distance between these representations. This is further passed through a linear layer (size: $2 \times 1$) along with the hypernym-to-hyponym score described in Section~\ref{select&substitute}. This linear layer is trained using a binary cross-entropy loss. 
    
    \item \textbf{SpanBERT} We use a SpanBERT model pretrained on SQuAD to find the answer from the retrieved snippets. We further fine-tune this model on \data by using the gold reformulated question, the snippets retrieved from it and the ground truth answers. 
    
    \item \textbf{\model classification} We merge the candidate answer representation (from the \textbf{\model} open-domain setting)  with the question representation using BLOCK's fusion module~\cite{ben2019block}. This fused representation is then fed to a softmax output layer with labels as the most frequent $k=1084$ answers for \data ($k=2048$ for \newdata) in the training set. The label with the highest score is returned as the predicted answer.
\end{description}

\subsection{Results}
\label{sec:ExptResults}

\paragraph{Classification Results.}

\begin{table}
\resizebox{\hsize}{!}{%
\begin{tabular}{cl|lll|}
\cline{3-5}
                       &  &   &    Reformulation  &  \\ \cline{3-5} 
                       &  & \multicolumn{1}{c|}{Original} & \multicolumn{1}{c|}{Gold} & \multicolumn{1}{c|}{Predicted} \\ \hline
\multicolumn{1}{|l|}{BLOCK} & W/o context & \multicolumn{1}{c|}{24.12} & \multicolumn{1}{c|}{25.74} & \multicolumn{1}{c|}{25.0} \\ \hline

\multicolumn{1}{|c|}{} & W/o MRC & \multicolumn{1}{c|}{12.03} & \multicolumn{1}{c|}{25.02} & \multicolumn{1}{c|}{19.44} \\ \cline{2-5} 

\multicolumn{1}{|c|}{\model} & With \mrcsingle{} & \multicolumn{1}{c|}{13.51} & \multicolumn{1}{c|}{31.81} & \multicolumn{1}{c|}{26.43} \\ \cline{2-5} 

\multicolumn{1}{|c|}{} & With \mrcmultiple{} & \multicolumn{1}{c|}{18.07} & \multicolumn{1}{c|}{33.55} &  \multicolumn{1}{c|}{28.57} \\ \hline

\end{tabular}  }
\caption{Classification results on \data. `W/o' expands as `without'.}
\label{table:Classification}
\end{table}

In Table~\ref{table:Classification}, we report classification results on \data using three different variants of \model. For these experiments, we use SpanBERT as our MRC system. The questions can either be in their original form (``Original"), or reformulated using the ground-truth annotations (``Gold") or reformulated using our span selector and substitution module (``Predicted"). We experiment with three variants of \model that use the top 10 snippets from Google concatenated together (henceforth referred to as \topten):
\begin{description}
    \item[Without MRC:] We bypass the use of an MRC module altogether; \topten is directly  fed as an input to the classification module. 
    \item[With \mrcsingle{}:] \topten is passed as input, along with the reformulated question, to SpanBERT. The output from SpanBERT, along with the reformulated question, are combined and fed as input to the classification module. 
    \item[With \mrcmultiple{}:] Each of the snippets in \topten are passed to SpanBERT to produce ten output spans. We use a simple attention layer over these spans to derive a single output representation, that is further fed as input to the classification module. 
\end{description}

The main observation from  Table~\ref{table:Classification} is that the MRC system, with its ability to extract relevant spans from the \topten snippets, is critical to performance. Using either \mrcsingle{} or \mrcmultiple{} in \model provides a substantial boost in performance compared to \model without any MRC system. And, \mrcmultiple consistently outperforms \mrcsingle. The BLOCK model significantly underperforms compared to \model.

\noindent {\bf Open-domain Results:}
In Table~\ref{table:opendomain}, we present open-domain results on both \data{} and our newly constructed \newdata{} using \mrcsingle. We report accuracies on all the three main steps of our proposed  architecture. While the open-domain setting is free of any answer bias, unlike the classification setting, it does not have the advantage of drawing from a fixed vocabulary of answers. Despite this perceived disadvantage, the best open-domain results on \data are comparable to (in fact slightly better than) the best classification results (i.e. 28.9 vs. 28.57). Recall from Section~\ref{sec:newdata} that unlike OKVQA, \newdata\ has exactly one correct ground truth answer per question leading to a much stricter evaluation and therefore much lower `search' numbers. Thus, while BLOCK yields close to $0$ evaluation score on \newdata (not reported in the table), results using \model{} are marginally better. 

\noindent {\bf Different MRC Systems.} We experiment with two different MRC systems, T5 and SpanBERT, and present open-domain results on \data using both these systems. The columns labeled ``without fine-tuning" refer to the original pretrained models for both systems and "with fine-tuning" refers to fine-tuning the pretrained models using \data. Fine-tuning the models with target data significantly helps performance, as has been shown in prior work~\cite{T5-finetune}. Scores labeled with (G) show the performance when select and substitute work perfectly.

\begin{table}
\begin{tabular}{ll|lll|}
 \cline{2-5} 
                       & \multicolumn{1}{|l|}{Test on} & \multicolumn{1}{l|}{Select} &  \multicolumn{1}{l|}{Substitute} &  Search\\ \hline\hline
\multicolumn{1}{|l|}{Train on \data\ } & \newdata & \multicolumn{1}{l|}{63.8} & \multicolumn{1}{l|}{24.6} & 14.6 \\
\cline{2-5} 
\multicolumn{1}{|l|}{Train on \newdata\ } & \newdata\ & \multicolumn{1}{l|}{94.9} & \multicolumn{1}{l|}{61.9} & 23.3  \\
\hline\hline
\multicolumn{1}{|l|}{Train on \newdata} & \data\ & \multicolumn{1}{l|}{57.2} & \multicolumn{1}{l|}{32.5} & 24.2 \\ \cline{2-5}
\multicolumn{1}{|l|}{Train on \data} & \data\ & \multicolumn{1}{l|}{67.2} & \multicolumn{1}{l|}{55.1} & 30.5 \\ \hline
\end{tabular}
\caption{Open-domain results on \newdata and \data. As for the low search results (marked *), we recall from Section~\ref{sec:newdata} that unlike OKVQA, \newdata\ has exactly one correct ground truth answer per question leading to a much stricter evaluation and therefore much lower numbers. In fact, BLOCK yields close to $0$ evaluation score on \newdata.} 
\label{table:opendomain}
\end{table}

\begin{table}
    \begin{center}
    {
        \begin{tabular}{|c|c|} 
            \hline
            MRC System & Testing on \data \\ 
            \hline\hline
            T5 & \begin{tabular}{c|c} without fine-tuning & with fine-tuning\\
            \hline  
                16 &  37.6 (G) 30.7 (P)
            \end{tabular} \\
            \hline
            SpanBERT & \begin{tabular}{c|c} without fine-tuning & with fine-tuning\\
            \hline  
                18 &  33.9 (G) 28.9 (P)
            \end{tabular} \\
        \hline
        \end{tabular}
    }
    \caption{Open-domain results using different MRC systems. (G) refers to ground-truth reformulated questions, (P) refers to reformulated questions predicted using \model.}
    \label{table:mrcs}
    \end{center}
\end{table}

%% file: main/sec7-conclusion.tex
\section{Conclusion}
In this paper, we identify key limitations of existing multimodal QA datasets in being opaque and uninterpretable in their reasoning. Towards addressing these limitations, we present \data, an improvisation on the existing OKVQA dataset as well as design and build a new challenge data set \newdata that focuses on a specific structural idiom that frequently appears in VQA. We also present a structurally transparent and interpretable system \model\  tailored to answer questions from our challenge data set and show that it outperforms strong baselines in both existing classification as well as the proposed open-domain settings.